\documentclass{article}

\usepackage{arxiv}

\usepackage[utf8]{inputenc} 
\usepackage[T1]{fontenc}    
\usepackage[hidelinks]{hyperref}       
\usepackage{url}            
\usepackage{booktabs}       
\usepackage{amsfonts}       
\usepackage{nicefrac}       
\usepackage{microtype}      
\usepackage{lipsum}
\usepackage{graphicx}
\graphicspath{ {./images/} }

\usepackage{amssymb}
\usepackage{tabularray}
\usepackage{color,soul}
\usepackage{changepage}
\usepackage[
    backend=biber, 
    style=ieee, 
    doi=true, 
    url=true, 
    isbn=true
]{biblatex}
\title{Multi-dataset synergistic in supervised learning to pre-label structural components in point clouds from shell construction scenes}

\author{
 Lukas Rauch \\
  Institute of Structural Engineering\\
  University of the Bundeswehr Munich\\
  Werner-Heisenberg-Weg 39, 85577, Neubiberg, Germany\\
  \texttt{lukas.rauch@unibw.de} \\
   \And
 Thomas Braml \\
  Institute of Structural Engineering\\
  University of the Bundeswehr Munich\\
  Werner-Heisenberg-Weg 39, 85577, Neubiberg, Germany \\
  \texttt{thomas.braml@unibw.de} \\
}

\begin{document}
\maketitle
\begin{abstract}
The significant effort required to annotate data for new training datasets hinders computer vision research and machine learning in the construction industry. This work explores adapting standard datasets and the latest transformer model architectures for point cloud semantic segmentation in the context of shell construction sites. Unlike common approaches focused on object segmentation of building interiors and furniture, this study addressed the challenges of segmenting complex structural components in Architecture, Engineering, and Construction (AEC). We establish a baseline through supervised training and a custom validation dataset, evaluate the cross-domain inference with large-scale indoor datasets, and utilize transfer learning to maximize segmentation performance with minimal new data. The findings indicate that with minimal fine-tuning, pre-trained transformer architectures offer an effective strategy for building component segmentation. Our results are promising for automating the annotation of new, previously unseen data when creating larger training resources and for the segmentation of frequently recurring objects.
\end{abstract}



\section{Introduction}

Spatial scene perception in computer vision has extensive use cases in the construction industry, helping to reduce tedious expert hours and enable autonomous robots on site. 
Most recent advancements in 3D perception, including segmentation and object detection, originated from research focused on self-driving cars and  logistics centers. Computer vision applications in the indoor built environment have predominantly targeted furnished residential spaces. This focus has resulted in a missed opportunity to extend these segmentation approaches to address common challenges on construction sites within the Architecture, Engineering, and Construction (AEC) domains.

At every stage of a building's life cycle, i.e., pre-construction, construction, renovation, and demolition, spatial perception with computer vision and the reconstruction of Building Information Models (BIM) have been shown to improve workflows, prevent critical errors, and avert costly and time-consuming rework. For instance, automating the classification of structural components can facilitate assembly progress control, subcontractor monitoring, and construction documentation. Additionally, generating an accurate 3D image can enhance plan precision while reducing planning efforts when building around existing structures. Furthermore, object detection and semantic map-building are crucial for the safe operation of autonomous machines on construction sites, especially in collaborative settings where detecting obstacles and people enhances safety for human workers and machines. 

The lack of publicly available data for training machine learning models presents a significant challenge for independent computer vision research in civil engineering, particularly in deep learning and the task of semantic component segmentation. Some recognized open-source benchmark datasets \cite{armeni.3D.2016,dai.ScanNet.2017a, zheng.Structured3D.2020} 
cover indoor spaces of residential and public buildings and include an intersecting set of classes with annotations for minor structural elements. These datasets typically offer limited variety beyond ceilings, floors, and walls. The creation of new domain-specific training data is essential, but data acquisition, curation, and, most of all, annotation are costly. Automatic pre-labeling can accelerate the annotation process for datasets derived from real-world data.

In this work, we address the challenges of semantic component segmentation for AEC in the context of a lack of training data by designing a series of experiments that leverage both existing benchmark datasets and a newly collected validation dataset specific to shell construction sites. 
We propose the hypothesis that established public datasets can be used to pre-label new, unseen data from construction sites across opposing domains. 

Our approach involved three main phases: (1) fully supervised training on our custom validation dataset, which served as a baseline for evaluating the model performance; (2) cross-domain training using multiple pre-existing datasets to assess the generalization capabilities of transformer-based segmentation models, and (3) transfer learning, where segmentation models were pre-trained on public indoor datasets and fine-tuned on our AEC specific data to enhance segmentation performance. The experiments were intentionally designed to explore the potential of combining domain-specific data with general datasets, to evaluate the effectiveness of different model architectures, and to determine the impact of pre-training on model performance. Through these experiments, we aimed to demonstrate how existing resources can be effectively utilized to improve semantic segmentation in the construction domain, ultimately bridging the gap between traditional computer vision approaches and the specialized needs of the AEC industry.

\section{Background}

\begin{figure*}[]
    \centering
    \includegraphics[width=1\textwidth]{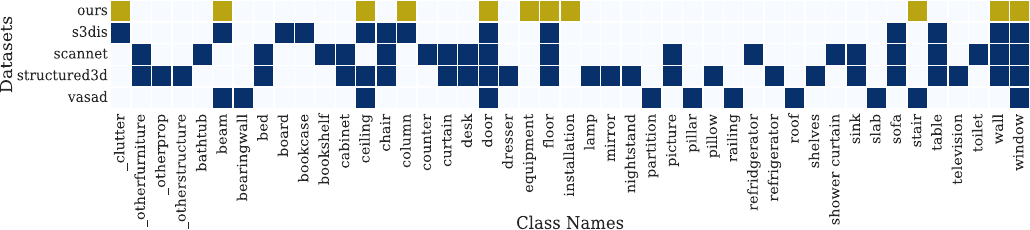}
    \caption{\textbf{One-hot encoding of the occurrence and overlap of object classes} of our validation dataset and four established training datasets S3DIS, ScanNet V2, Structured3D, and VASAD. The listed datasets are used by us for training models for the semantic segmentation of component groups.}
    \label{fig:class_encoding}
\end{figure*}

Semantic segmentation is a task in computer vision that involves classifying each pixel in an image or each point in a point cloud into predefined categories, effectively partitioning the entire scene into meaningful segments. Unlike traditional classification tasks, which assign a single label to an entire image or object, semantic segmentation provides a more detailed understanding by labeling every element within the scene. Supervised deep learning requires large amounts of data, which are not always readily available. Therefore, innovative approaches have been developed to maximize the utility of existing datasets. The following paragraphs briefly summarize the advances and milestones in deep learning-based point cloud semantic segmentation, which form the foundation for this work.

\subsection{General Point Cloud Semantic Segmentation}
Most approaches to point-based semantic segmentation refer to the 2017 released PointNet \cite{qi.PointNet.2016a}. A detailed summary of the key milestones that followed in subsequent years is beyond the scope of this paper; therefore, the authors recommend the survey by Guo, et al. \cite{guo.Deep.2021} for a comprehensive review of the fundamentals and developments in deep learning for point clouds and 3D semantic segmentation up to the end of 2020. The comparison of various benchmark results revealed that in recent years, transformer-based deep learning architectures have outperformed their counterparts (Point-CNN, Graph-CNN, RNN) in most point cloud perception tasks \cite{wu.Point.2022, yang.Swin3D.2023, paperswithcode.3D.2024, rauch.Semantic.2023}. These point transformers were inspired by the success of transformers in natural language processing (NLP) and 2D image vision.

The core of all transformer models is a self-attention mechanism that is invariant to the permutation and cardinality of the input elements. Applying self-attention to 3D point clouds is intuitive since point clouds are essentially sets embedded irregularly in a metric space \cite{zhao.Point.2021}. The naïve transformer computes global self-attention across the entire point cloud, enabling long-range attention between scattered point patches \cite{guo.PCT.2021}. However, this approach leads to high memory and computational costs due to the quadratic complexity of self-attention \cite{wu.Point.2022}.

Zhao et al. \cite{zhao.Point.2021} introduced local attention around each data point (using k nearest neighbors), which reduced complexity and made Point Transformers feasible at the scene level. Wu et al.’s Point Transformer V2 \cite{wu.Point.2022} employed grouped vector attention, allowing the creation of deeper networks, enhanced position encoding, and partition-based pooling to address irregular spatial point distribution. The model’s generalization capabilities were further improved with a wider receptive field, which Wu et al. \cite{wu.Point.2023} achieved by prioritizing simplicity and efficiency. Around the same time, the Stratified Transformer \cite{lai.Stratified.2022} adopted a grid-based sliding window attention mechanism from 2D vision’s Swin Transformer \cite{liu.Swin.2021}, enabling transformer blocks to operate within a sequence of shifted windows on a 3D voxelized point cloud. Swin3D \cite{yang.Swin3D.2023} enhanced the naive window attention for sparse 3D voxel grids by reimplementing multi-head self-attention. This reduced memory costs from quadratic to linear with respect to the number of sparse voxels per window, allowing for a wider receptive field and improved generalization.

\subsection{Pre-Trained Transformer Backbones}
The integration of large pre-trained backbones has driven significant advancements in the NLP and 2D vision domains, enabling better task generalization, streamlined network design, and reduced requirements for labeled data and training time \cite{liu.Swin.2021, bao.BEiT.2021, devlin.BERT.2018}. This approach involved pre-training a general backbone network on broad data, which could then be fine-tuned for various downstream tasks, such as segmentation or object detection. Swin3D \cite{yang.Swin3D.2023} proposed a pre-trained transformer backbone for general indoor 3D scene understanding. The model was trained on the synthetic dataset Structured3D \cite{zheng.Structured3D.2020} and can be further fine-tuned for downstream tasks. Experiments on point cloud semantic segmentation demonstrate robust domain generalization capabilities across multiple real-world datasets, highlighting the particular advantages of transfer learning for small datasets.

\subsection{Point Cloud Segmentation for the AEC} 
Observing research over the past few years, it became evident that significant contributions from the community to industry-specific segmentation approaches are primarily focused on the architecture aspect of AEC (Architecture, Engineering, and Construction). Developing deep learning-based semantic segmentation methods for indoor scene context largely relies on the major benchmark datasets like S3DIS \cite{armeni.3D.2016}, ScanNet v2 \cite{dai.ScanNet.2017a}, and Structured3D \cite{zheng.Structured3D.2020}. However, these datasets primarily feature interior design and furniture classes alongside basic building elements such as ceilings, floors, walls, beams, doors, and windows. The application of such segmentation systems to engineering disciplines and construction may require a more technical understanding and, thus, a broader range of represented component classes.

In addition to publications that aim for top performance on the three significant benchmarks (S3DIS, ScanNet v2, and Structured3D), several notable papers have influenced the research on point cloud semantic segmentation from an AEC perspective. The VASAD dataset \cite{langlois.VASAD.2022} took a synthetic data approach to segmentation-based scene reconstruction in the scope of structural engineering. It consists of six digital computer-aided design (CAD) building models with full-volume descriptions and semantic labels. Synthetic training data can be collected by raytracing virtual laser beams within these CAD models, which allows the generation of theoretically infinite scans from virtually any camera position. For our work, we used the pre-rendered point clouds  by the VASAD authors, which contained a set of 11 ground truth semantic labels, all considered construction-related. The occurrence of the object classes and their common overlap among the four other datasets is shown as one-hot encoding in Figure \ref{fig:class_encoding}.

Ma et al. \cite{ma.Semantic.2020a} explored the effectiveness of augmenting S3DIS training data with synthetic data from 3D BIM models, while Noichl et al. \cite{noichl.Enhancing.2024} tackled similar challenges in the environment of industrial plants. Additionally, several papers have addressed the segmentation of Mechanical, Electrical, and Plumbing (MEP) installations on small-scale personal datasets using heuristic methods, such as feature-based region growing for geometric shape recognition \cite{dimitrov.Segmentation.2015, yin.Automated.2021a} and a convolutional, residual point cloud feature learning approach \cite{perez-perez.Segmentation.2021a}. However, the limited availability of high-quality training and validation data remains one of the most significant barriers to advancing semantic segmentation for the AEC industry.

\subsection{Multi-Dataset Synergistic Training} 
The strategy of merging multiple data sources to train a single model collaboratively has shown promising results in scenarios with limited training resources \cite{wu.Largescale.2023}. Combining similar datasets to enrich the training pool has proven beneficial in 2D scene understanding with studies \cite{wang.CrossDataset.2021, kim.Learning.2022} reporting improved model generalization on unseen datasets, even when label spaces differ. However, in the 3D domain, the significant domain gap and sparse nature of datasets can lead to negative transfer, potentially harming model performance when naively combined. 

\section{The Validation Dataset}

\begin{figure*}[]
  \centering
  \includegraphics[width=1\textwidth]{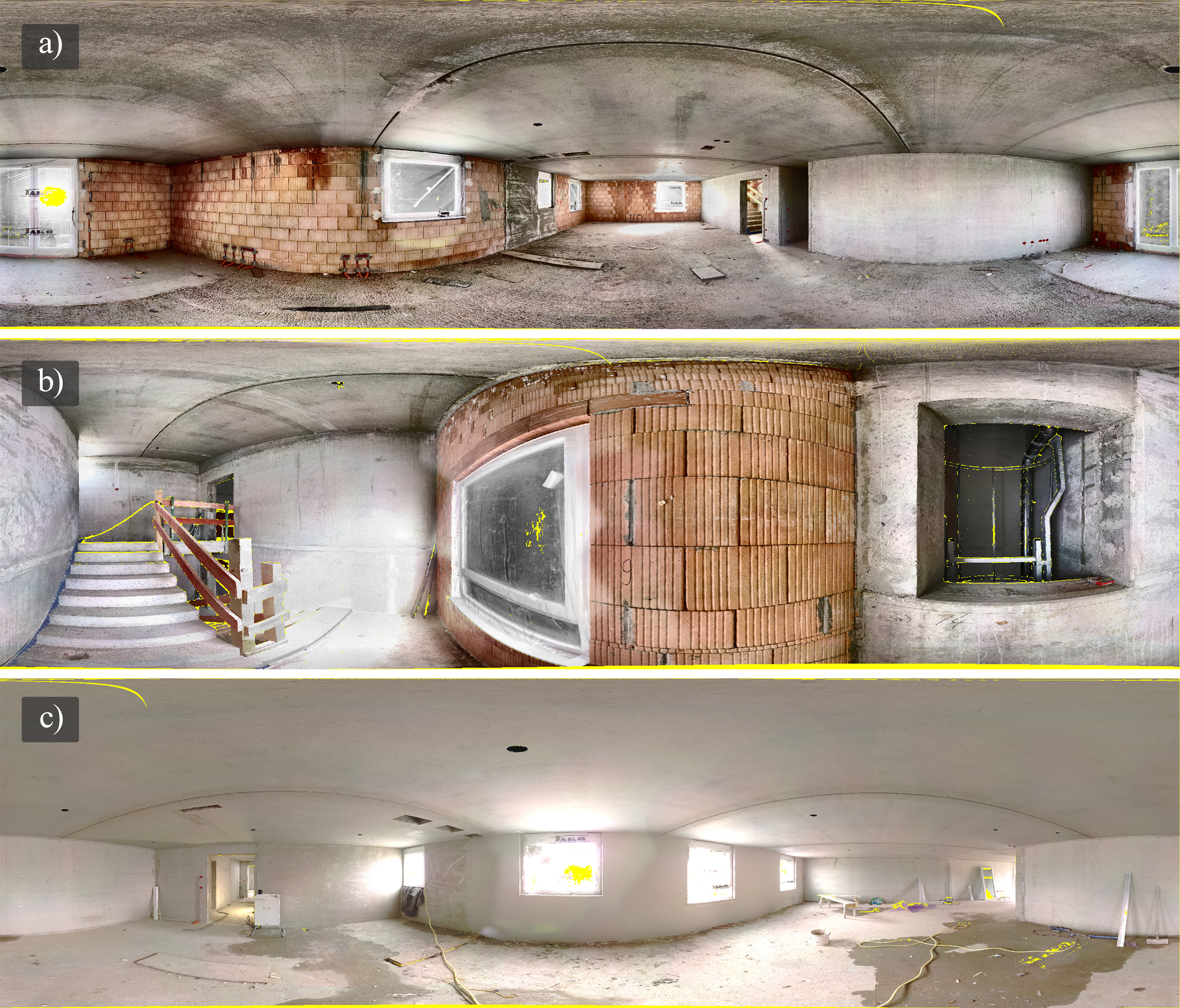}
  \caption{\label{fig:rb3d_panorama} \textbf{Spherical Point Cloud Renderings} of three rooms from the custom validation dataset, collected at a residential apartment building site during shell construction. \textbf{a)} Medium-sized room before plastering work is completed. \textbf{b)} Staircase before plastering work is completed. \textbf{c)} Medium-sized room after plastering work is completed. The scenes are characterized by varying surface textures due to the nature of the construction, challenging lighting conditions, and complex floor plans. They include obstacles and wet spots on the floor, which produce reflections and scanning artifacts. Yellow pixels represent empty canvas pixels where no points were projected.}
\end{figure*}

In this study, we created a test dataset featuring diverse indoor scenes from a shell construction site for validation purposes. Three spherical RGB point cloud-to-image renderings of representative scenes are shown in Figure \ref{fig:rb3d_panorama}. The dataset was collected using a FARO Focus M70 terrestrial laser scanner, a device commonly used in construction management and supervision to measure, document, and evaluate ongoing construction processes. Compared to mobile LiDAR systems, the terrestrial scanner produces a very high point density, high 3D accuracy, and a low noise range.

The presented dataset comprises 36 scenes from a multi-story residential building, including various apartment floor plans and stages of completion. The combined point clouds consist of approximately $3.6 \times 10^8$ points, with a mean neighborhood point density of 6.67 mm (standard deviation = 0.002 mm), calculated based on the 30 nearest neighboring points. The scans were collected over the course of an entire day, capturing naturally varying daylight conditions. The data includes per-point RGB colors, the laser's surface reflection strength, and ground truth annotations at the class and instance levels provided by domain experts.

The scenes in the dataset vary in lighting conditions, surface finishes (e.g., brick walls, concrete, plaster), and levels of contamination (e.g., interfering elements, rubble). As is typical for laser scans and how the device perceives the environment, the dataset contains various challenging details, such as reflection artifacts from mirroring surfaces (e.g., wet spots) and transparent objects (e.g., glass windows). The individual scans are not registered, and for each point set, the scanner's center position is located at the coordinate origin. 
The geographical orientation around the z-axis in individual scenes does not follow a fixed, cardinal direction, appearing quasi-random as output from the device.

Point cloud preprocessing was kept to a minimum. The only manipulation of the raw data involved applying an Euclidean distance filter to exclude points more than 25 meters from the coordinate origin. This step reduces the number of faulty measurements, often caused by highly reflective surfaces or other sensor disturbances, which otherwise add no value to the analysis and may even corrupt it.

A single specialist annotated the ground truth labels to ensure consistency across the entire dataset. Unit-based annotation means that each individual scan point was assigned to one of 11 unique object classes. These classes include \textit{ceiling, floor, wall, beam, column, window, door, stairs, equipment, installation}, and a generic \textit{none} class for anything that cannot be classified. This list represented a selection of the most frequently recurring component groups in building construction and was assumed to provide a good generalization.

\begin{figure*}[]
    \centering
    \includegraphics[width=\textwidth]{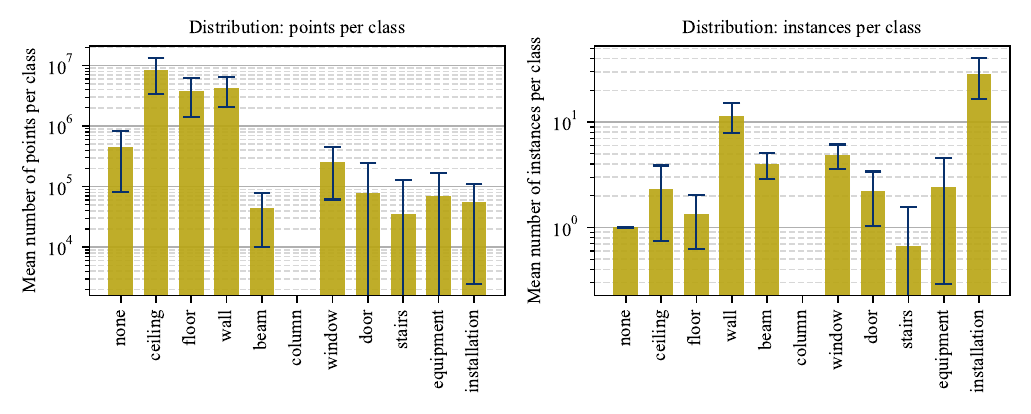}
    \caption{\textbf{Statistical Evaluation} of the mean points-per-class distribution and mean instances-per-class distribution per scene in the validation dataset, plotted on a logarithmic scale. The uncertainty bars represent the standard deviation of the statistical sample distribution between the 36 scenes.}

    \label{fig:rb3d_class_distribution}
\end{figure*}

Figure \ref{fig:rb3d_class_distribution} provides a statistical evaluation of the points-per-class distribution and instances-per-class distribution within the validation dataset. This representation helps visualize which classes hold how many points or instances and assess the dataset's internal balance. The \textit{wall, floor}, and \textit{ceiling} classes, which constitute the majority of interior surfaces, contain the majority of all individual points (94.3\%) in the points-per-class view. Correspondingly, these large-area components appear less frequently in the instances-per-class view. Small components, such as electrical installation shafts, have many instances but are represented by relatively few scan points. Rare component groups, such as stairs and beams, tend to be underrepresented. The \textit{column} class is not at all represented in this small validation dataset since all 36 samples were collected from residential buildings where columns are uncommon. At this stage, the class serves as a placeholder for future experiments with more diverse construction sites.
The authors acknowledge that this may impact certain averaging validation metrics, such as the mean intersection over union (mIoU). Therefore, the evaluations in Section \ref{sec:results} primarily consider the classes individually, providing deeper insights into the model's confusion and misclassification between the classes rather than quantifying performance solely by an aggregated metric.

It is recommended that the dataset be split into partitions at a ratio of 70/15/15 for training, validation, and testing in supervised machine learning. We carefully selected the validation and test splits to ensure they contain genuinely unseen data, as multiple samples in the dataset may share common content from overlapping scans. The details of the dataset split used in the experiments are provided in Table \ref{tab:datasplit}.

\def\tabcolwidth{.7cm}
\begin{table}[!h]
    \caption{\textbf{Data Split for the custom validation dataset} used for all experiments in this paper. The 36 samples are split in a ratio of 70\%/15\%/15\%.}
    \label{tab:datasplit}
    \centering
    
    \begin{tblr}{
    Q[l,f,1.5cm] Q[c,f,5cm] Q[c,f,2cm]} 

    \textbf{Task} & \textbf{Sample Selection} & \textbf{Total Sum} \\
    \hline

        Training    & 0-11, 14, 15, 20-23, 30-35  & 24 \\
        Validation	& 16-19, 28, 29	              & 6 \\
        Testing	    & 12, 13, 24-27               & 6 \\

    \end{tblr}
\end{table}

\section{Experimental Design}

The objective of the experiments conducted in this work was to test the extent and the quality to which established datasets can be used to pre-label new point cloud data with classes related to shell construction. The target classes were a selection of the most frequently occurring structural component groups and objects in building design and shell construction. These 11 classes included \textit{ceiling, floor, wall, beam, column, window, door, stair, equipment, installation}, and a \textit{none} class to account for clutter, noise, and any non-construction-related objects that may have been present in the scene.

No available dataset contained all the required classes nor a large intersection of them. Therefore, we utilized three common training resources for 3D semantic segmentation in indoor environments: S3DIS \cite{armeni.3D.2016}, ScanNet v2 (20) \cite{dai.ScanNet.2017a}, and Structured3D \cite{zheng.Structured3D.2020},  and compared them in a multi-dataset synergistic approach to cover as much context as possible. We also included the smaller VASAD dataset \cite{langlois.VASAD.2022} due to its larger intersection with our target classes. These four datasets are referred to as \textit{baseline datasets} in the following sections. The actual label overlap between the baseline datasets and our reference dataset is shown in Figure \ref{fig:class_encoding}. The experiments were divided into three steps to test the hypothesis that established public datasets could be used to pre-label new, unseen data from different domains.

\subsection{Baseline Training}

The baseline for our experiments was established by training the three model architectures Point Transformer V2 \cite{wu.Point.2022}, Point Transformer V3 \cite{wu.Point.2023}, and SWIN3D (small) \cite{yang.Swin3D.2023} on our test dataset in a fully supervised manner. 
The training split consisted of only 24 scenes, a small data set for a general deep learning approach, but this represented roughly the amount of data that could be provided in practice with reasonable effort on a scene-by-scene basis.  
The segmentation results from this test served as a low bar reference that the following two approaches had to surpass to confirm our hypothesis. We compared the performance of three architectures and focused on their generalization ability from a small training set. 
The semantic segmentation performance could be quantified in numbers and compared using the metrics mean accuracy (mAcc) and mean intersection over union (mIoU). Accuracy is the ratio of correctly classified points to the total number of points per class.
Intersection over union quantifies the overlap between the predicted segmentation region and the ground truth annotated region from a reference dataset. The averaging mIoU provides a balanced view of the model's performance, particularly in scenarios where the dataset has an uneven distribution of classes, which was the case with the validation dataset.

\subsection{Cross-Domain Evaluation}

\def\tabcolwidth{1.2cm}
\begin{table}[]
    \caption{\textbf{Dataset - model pairings} that were evaluated in the cross-domain experiment.}
    \label{tab:cross_domain_combinations}
    \centering
    \begin{tblr}{Q[l,f,1.6cm] Q[c,f,\tabcolwidth] Q[c,f,\tabcolwidth] Q[c,f,\tabcolwidth] Q[c,f,\tabcolwidth]} 

     & ScanNet & S3DIS & Struct.3d & VASAD \\
    \cline{2-5}
    PVv2    &   &  $\checkmark$ &  $\checkmark$ &    \\
    PVv3    & $\checkmark$ &  $\checkmark$ &  $\checkmark$ & $\checkmark$  \\
    SWIN3D  &   &  $\checkmark$ &  $\checkmark$ &    \\
    
    \end{tblr}
\end{table}

We trained eight combinations of three model architectures across four baseline datasets. The dataset - model pairings are presented in Table \ref{tab:cross_domain_combinations}. We evaluated each combination against our custom validation data to assess their generalization capabilities toward the domain shift. The final segmentation layer of the network remained unchanged during this step, resulting in predictions aligned with the class notation of the baseline datasets, as illustrated in Figure \ref{fig:class_encoding}. As a closed-set classifier, each network could only predict the classes for which it was trained, leading to a mismatch between the predicted classifications and the annotations of our validation dataset. Depending on the baseline dataset used for training, the predicted class IDs ranged from 10 possible classes for models trained on VASAD to 25 possible classes for those trained on Structured3D. The validation targets include 11 classes. 

To calculate the performance metrics post-inference, the ground truth data label set had to match the label set of the model predictions. However, this condition is violated when models are trained and tested on data with different label sets. To reconcile this mismatch, we introduced an additional translation layer at data loading time that adapted the order of the validation labels to align with the label indices of the respective training dataset. 
This translation followed the class overlap depicted in Figure \ref{fig:class_encoding}. 
Classes from the validation dataset that were not present in the respective training dataset were assigned to one of the \textit{\_none}-classes (indicated by an underscore). They were excluded from the calculation of the performance metrics afterward. 
The results obtained from this approach were used both to assess the synergy between the baseline training datasets and the validation dataset and to determine which dataset-model pairings were suitable for further fine-tuning experiments. 

Due to the poor segmentation performance observed in the first two experiments, as detailed in the results and discussion section, we excluded the underperforming datasets, ScanNet and VASAD, and the PTv2 architecture from any subsequent fine-tuning steps. Our definition of poor performance in the experiments was a fundamental failure of the models to understand even the primary groups of components (\textit{ceiling, floor, wall}), which we assessed using quantitative and visual results.

\subsection{Transfer Learning}

\begin{figure*}[!h]
    \centering
    \includegraphics[width=\textwidth]{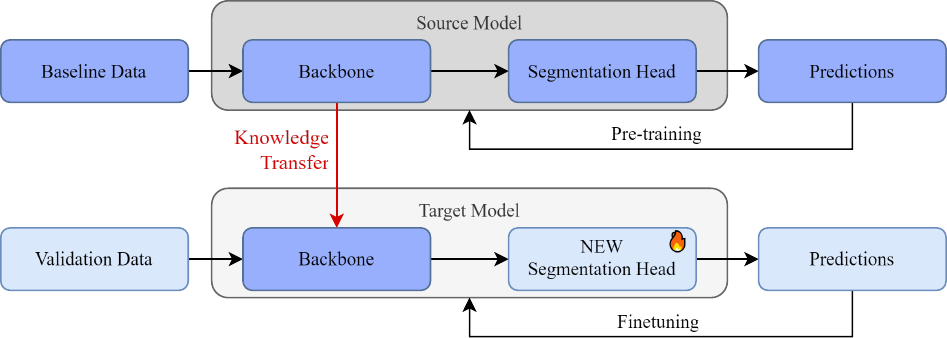}
    \caption{\textbf{Concept of transfer learning}, in which the knowledge from the previous training is repurposed to improve the downstream tasks. }

    \label{fig:transferlearning_scheme}
\end{figure*}

We utilized the two models, Point Transformer V3 and SWIN3D (small), which had been pre-trained in the previous step with the S3DIS and Structured3D datasets, and further fine-tuned them with our custom validation data using a transfer learning approach. An abstract representation of the basic transfer learning concept is shown in Figure \ref{fig:transferlearning_scheme}. 
The source model was originally trained on the class labels of the baseline dataset. Simply put, the basic framework of the model was responsible for learning 3D features, and in the specific downstream task of semantic segmentation, a dedicated segmentation head generated the point-based class predictions.
This segmentation head was bound to the defined classes of the baseline dataset and could therefore only recognize those classes. The backbone, on the other hand, learned label-independent spatial perceptual features during pre-training. This prior knowledge was transferred to a target model, where it was fine-tuned for the target classes using a small amount of validation data. 
The segmentation head, responsible for point-wise classification based on the pre-training data classes, was reinitialized with random weights and biases. This adjustment involved replacing the segmentation head at the end of the neural network with a new segmentation head configured for output dimensions required for validation. 
During the fine-tuning steps, the model is expected to utilize the previously learned information contained in its backbone to reduce the required training data and improve the overall segmentation results.
The backbone network was not frozen during training and could slightly adapt to the new data. To avoid drastic changes, we reduced the maximum value of the scheduled OneCycleLearningRate hyperparameter considerably from 0.006 to 0.001. 

\subsection{General Experiment Configurations}

For all experiments in this work, we used the open-source Pointcept codebase \cite{pointceptcontributors.Pointcept.2023}, which provides implementations of multiple state-of-the-art point transformer models. An extensive literature review \cite{rauch.Semantic.2023} showed that, in recent years, transformer architectures had emerged as superior benchmarks for 3D semantic segmentation on all relevant indoor point cloud datasets \cite{paperswithcode.3D.2024}. We implemented a custom data loader, a label translation layer, and the necessary pipeline to train and evaluate our reference dataset within this codebase. The original code extensions can be found on the author’s GitHub\footnote{To be added after final revision}.

\begin{figure*}[]
  \centering
  \includegraphics[width=\textwidth]{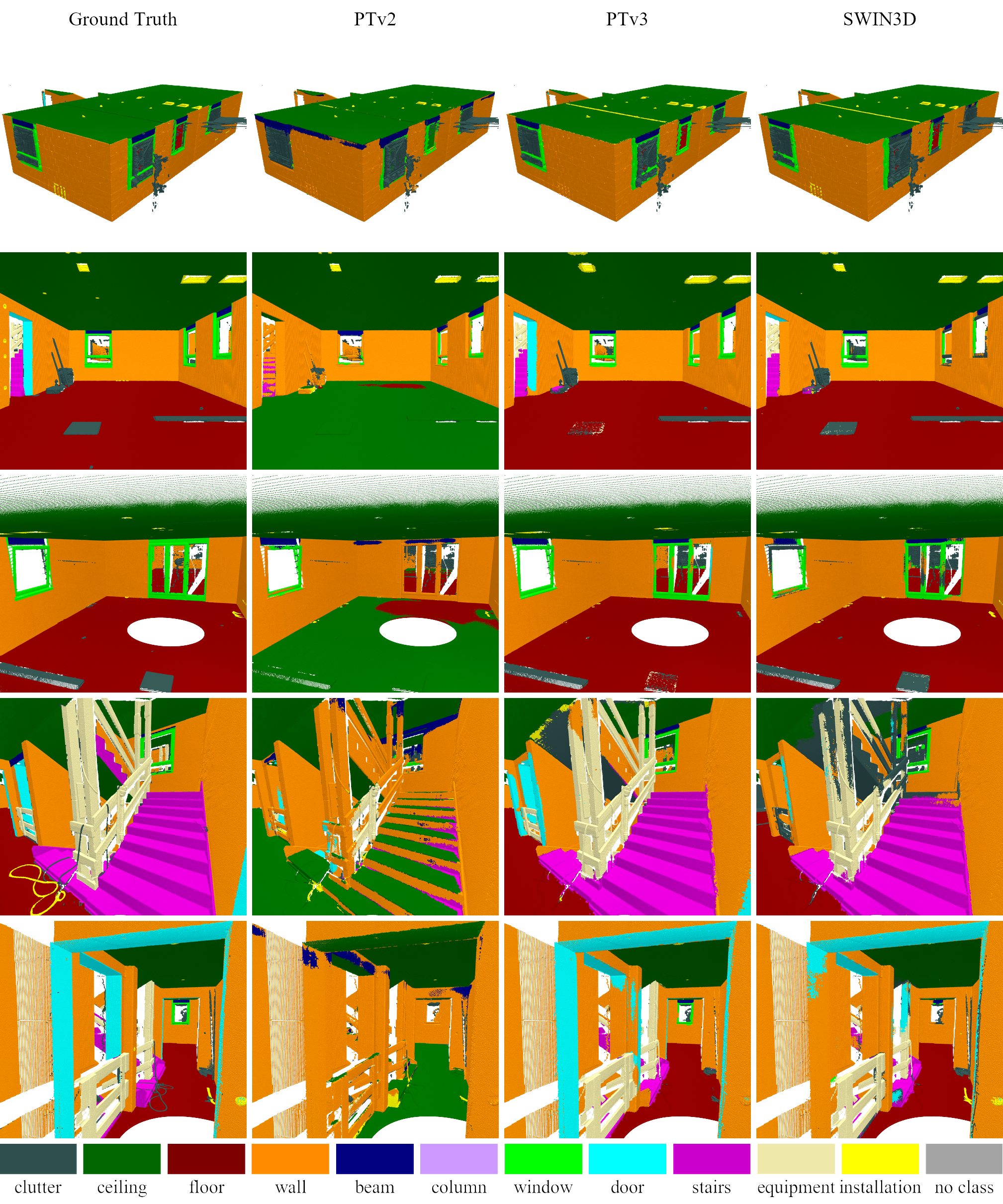}
  \caption{\label{fig:baseline} \textbf{Baseline Test Results for 3D Semantic Segmentation.} 
  This figure presents the inference results from the baseline training experiment using three model architectures: Point Transformer V2, Point Transformer V3, and SWIN3D. Each model was trained and tested on our custom validation dataset, which focuses on shell construction site scenes. The columns display five representative screenshots per model, with predicted class labels uniquely colored to illustrate the segmentation performance across different architectural components.}
\end{figure*}

During training, model evaluation was performed on a single grid-based subsample of the point cloud, providing an initial assessment of model performance. The precise testing process involved grid-sampling a dense point cloud into a sequence of point cloud fragments to ensure comprehensive coverage of all points. Segments were then predicted, aggregated for all fragments, and assigned to the points by class voting, forming a complete prediction of the entire point cloud. This approach yielded higher evaluation results than simply mapping or interpolating the prediction from the subsampled point cloud to its original shape. 
The Pointcept library \cite{pointceptcontributors.Pointcept.2023} also allowed us to use data augmentation during training to prevent overfitting and test-time augmentation (TTA) \cite{kimura.Understanding.2021} during the final precise evaluation step to further enhance the stability of the evaluation performance. 
Test-time augmentation creates instances of the original test sample but with augmenting transformations applied upfront of the grid-sampling fragmentation. 
Quantitative performance results in Table \ref{tab:rb3d_baseline_miou}, Table \ref{tab:transfer}, and Table \ref{tab:finetune} that utilized the enhanced evaluation are marked with an asterisk (*). All visualized results in Figure \ref{fig:baseline}, Figure \ref{fig:transfer_1}, Figure \ref{fig:transfer_2}, and Figure \ref{fig:finetune} were produced using testing through TTA enhancement.

All baseline models were trained from scratch using equivalent training hyperparameters to ensure consistency across the experiments. Our validation dataset did not include surface normal vectors as point cloud features. Surface normals for point clouds are typically derived from post-processing or are a convenient by-product of creating synthetic datasets. Still, they are not directly obtainable from the laser scanner device. Therefore, the corresponding normal features from the S3DIS and Structured3D datasets were omitted in this comparison, and the models were trained with coordinates and point colors as the only input features. The VASAD experiments were performed exclusively with coordinates as input features, as this synthetic dataset did not contain colored point clouds.

The hyperparameters were derived from empirical values and recommendations in the Pointcept codebase \cite{pointceptcontributors.Pointcept.2023}, without dedicated optimization by the authors. The training was performed on multiple Nvidia Tesla V100 SXM3 graphics cards with 32 GB of video memory. An overview of the hyperparameters and training configurations is provided in Table \ref{tab:hyperparameters}.

\begin{table}[]
    \caption{\textbf{Overview of the hyperparameters} in the experiments for training.}
    \label{tab:hyperparameters}
    \centering

    \begin{tblr}{
        colspec={Q[l,h,0.12\textwidth] Q[m,l,0.12\textwidth] Q[m,l,0.65\textwidth]},
        rowsep=6pt 
    } 
    \hline

        \multirow[t]{4}{0.12\textwidth}{Training Configuration} & Loss function & The sum of Cross Entropy loss and Lovász-Softmax loss \cite{berman.LovaszSoftmax.2018}. \\ \cline{2-3}
        
         & Optimizer & AdamW, scheduled with a one-cycle learning rate and a decent cosine. \\ \cline{2-3}

         & Epochs & Depending on the convergence speed of the model/dataset pairing. Early stop if loss converges.  \\ \cline{2-3}

         & Batch Size & 2 to 8 batches per card. Depending on the sample size, which again depends on the dataset, its point density, and the number of model parameters. The feasible batch size is restricted by the limited graphic card memory.\\ \hline
         
        \multirow[t]{2}{0.12\textwidth}{Data \\ Augmentation} & Geometric Space & Center shift, random dropout, random rotate, random flip, random jitter, voxelization (voxel size 0.025 m), random sphere crop ($\sim$100K points). \\ \cline{2-3}
        
        & Feature Space & Chromatic auto contrast, chromatic translation, chromatic jitter, normalize color. \\ 
    \end{tblr}

\end{table}

\section{Results and Discussion}
\label{sec:results}

In this section, we evaluate and discuss the test results of the initial experiments in relation to our research hypothesis. Our hypothesis posits that a model can be effectively trained with minimal new data, thereby significantly simplifying the pre-labeling process for data annotation. The utility of the transformer models for this purpose was assessed through a series of experiments following our experimental design.

\def\tabcolwidth{1.2cm}
\begin{table}[]
    \caption{\textbf{Baseline Validation and Test Results for 3D Semantic Segmentation} 
     for three model architectures — Point Transformer V2, Point Transformer V3, and SWIN3D — trained on our validation dataset for the task of 3D semantic segmentation. The performance metrics include mean Intersection over Union (mIoU), mean Accuracy (mAcc), and overall Accuracy (allAcc), which are reported for both validation during training and final testing. The test results marked with an asterisk (*) indicate the use of Test Time Augmentation (TTA) in the final testing to enhance prediction stability.}
    \label{tab:rb3d_baseline_miou}
    \centering
    \begin{tblr}{Q[l,f,\tabcolwidth] Q[c,f,\tabcolwidth] Q[c,f,\tabcolwidth] Q[c,f,\tabcolwidth] Q[c,f,\tabcolwidth] Q[c,f,\tabcolwidth]Q[c,f,\tabcolwidth]} 

    \textbf{Archit.} & \textbf{mIoU} & \textbf{mAcc} & \textbf{allAcc} & \textbf{mIoU*} & \textbf{mAcc*} & \textbf{allAcc*} \\
    \hline

    PTv2 &   .67 & .75 & .97	& .29 & .41 & .75 \\
    PTv3  &  .61 & .72 & .95	& .66 & .77 & .97 \\
    SWIN3D & .50 &	.57 & .93 & .60 & .72 & .97 \\
    
    \end{tblr}
\end{table}

\def\tabcolwidth{.6cm}
\begin{table}[]
    \caption{\textbf{Baseline Test Results for 3D Semantic Segmentation} 
    for three different model architectures — Point Transformer V2, Point Transformer V3, and SWIN3D — trained on our validation dataset for the task of 3D semantic segmentation. The performance metrics include class-wise Intersection over Union (IoU) and Accuracy (Acc), which were reported for validation during training.}
    \label{tab:rb3d_baseline_all_iou}
    \centering
    \begin{tblr}{Q[l,f,2.5cm] Q[c,f,\tabcolwidth] Q[c,f,\tabcolwidth] Q[c,f,\tabcolwidth] Q[c,f,\tabcolwidth] Q[c,f,\tabcolwidth]Q[c,f,\tabcolwidth]} 

     & \SetCell[c=2]{f,c} PTv2 & & \SetCell[c=2]{f,c} PTv3  & & \SetCell[c=2]{f,c} SWIN3D \\

    \textbf{target class} 
      & \textbf{IoU} & \textbf{Acc} 
      & \textbf{IoU} & \textbf{Acc}
      & \textbf{IoU} & \textbf{Acc} \\
    \hline
    \_none       & .50 & .60 & .69 & .76 & .60 & .85 \\
    ceiling      & .69 & 1.0 & .97 & .97 & .98 & .98 \\
    floor        & .05 & .05 & .98 & 1.0 & .99 & 1.0 \\
    wall         & .89 & .98 & .97 & .99 & .95 & .98 \\
    beam         & .27 & .97 & .57 & .96 & .41 & .70 \\
    column       & 0   & 0   & 0   & 0   & 0   & 0   \\
    window       & .19 & .19 & .71 & .80 & .55 & .61 \\
    door         & .07 & .07 & .81 & .87 & .56 & .58 \\
    stair        & .14 & .14 & .84 & .93 & .86 & .90 \\
    equipment    & .07 & .07 & .61 & .64 & .58 & .67 \\
    installation & .35 & .44 & .07 & .52 & .14 & .61 \\
    
    \end{tblr}
\end{table}

\subsection{Baseline Results}

The quantitative results in Table \ref{tab:rb3d_baseline_miou} revealed that both PTv3 and SWIN3D delivered mixed outcomes for semantic segmentation, with a mean Intersection over Union (mIoU) in the lower 60\% range. The average value alone was of limited use; a deeper insight into the individual classes, presented in Table \ref{tab:rb3d_baseline_all_iou}, showed that the IoU score was predominantly influenced by three classes: \textit{wall, ceiling}, and \textit{floor}, which are primarily planar surface elements. These classes achieved an average IoU of 98\%. \textit{Stairs} could be segmented in the point cloud with an IoU above 80\% by both PTv3 and SWIN3D, demonstrating their ability to handle specific non-planar elements effectively.
A practical use case for this and the following two experiments would be a point cloud preprocessing filter to clean raw sensor data from outliers and clutter. However, the achieved \textit{none}-class IoU of about 60\% in this experiment is still too low for such an application where a low false positive rate is required. In total numbers, PTv3 achieved a slightly higher median IoU (71\%) than SWIN3D (58\%). This suggests that the Point Transformer V3 model might be better suited for segmenting minority classes that occur less frequently in the small training data pool.
Conversely, the PTv2 model failed for undetermined reasons. Validation and testing results (shown in Table \ref{tab:rb3d_baseline_miou}) were inconsistent, with the \textit{floor} class not recognized at all during precise testing (indicated by an asterisk *) and other classes performing significantly worse than in the PTv3 model. This discrepancy was unexpected given that the differences between PTv2 and PTv3 primarily relate to simplicity, efficiency, and scalability rather than accuracy \cite{wu.Point.2023}. During training, PTv2 performed best and outperformed the validation metrics of both PTv3 and SWIN3D. Yet, this performance was not reproducible with the test dataset during inference. A plausible explanation is that PTv2 has a slightly higher model complexity than PTv3, leading to overfitting on the small training dataset and an inability to generalize to new data. Consequently, we decided to exclude PTv2 from subsequent experiments, as its successor, PTv3, provided higher segmentation results with lower computational costs.
The qualitative results in Figure \ref{fig:baseline} corroborated the quantitative findings. Both PTv3 and SWIN3D achieved consistent segmentation results for \textit{ceiling, floor}, and \textit{wall} elements. Visual inspection indicated that SWIN3D was more adept at segmenting small elements, such as electric \textit{installations} and \textit{clutter}, which frequently occur within the training dataset but are often tightly embedded in the \textit{ceiling, floor}, and \textit{walls}. In contrast, PTv3 excelled in segmenting door frames and windows, even in challenging scenarios where some windows were opened and tilted from their original orientation. Both models struggled to segment the story-high window, which deviates from the typical window shape.
The current results, while promising, are not yet sufficient for practical application. Smaller component objects must be segmented more reliably to achieve a final application-ready model. Nonetheless, these findings support our initial thesis that a relatively small dataset can effectively pre-label larger datasets.

It's important to note that these experiments were conducted on a very small-scale dataset. Despite no overlap of the individual samples and the use of unseen building sections for validation and testing, all validation data comes from a single large building complex. Therefore, the results may not generalize well to other buildings, even those of the same building type. However, the experiment remains relevant if we consider labeling a small subset of samples by hand for each building to be annotated and using this set to train a model for pre-labeling the remaining samples. Even if the results are insufficient for in-production segmentation applications, they may still help obtain class boundary boxes when precise segmentation is not required.

\begin{figure*}[]
  \centering
  \includegraphics[width=\textwidth]{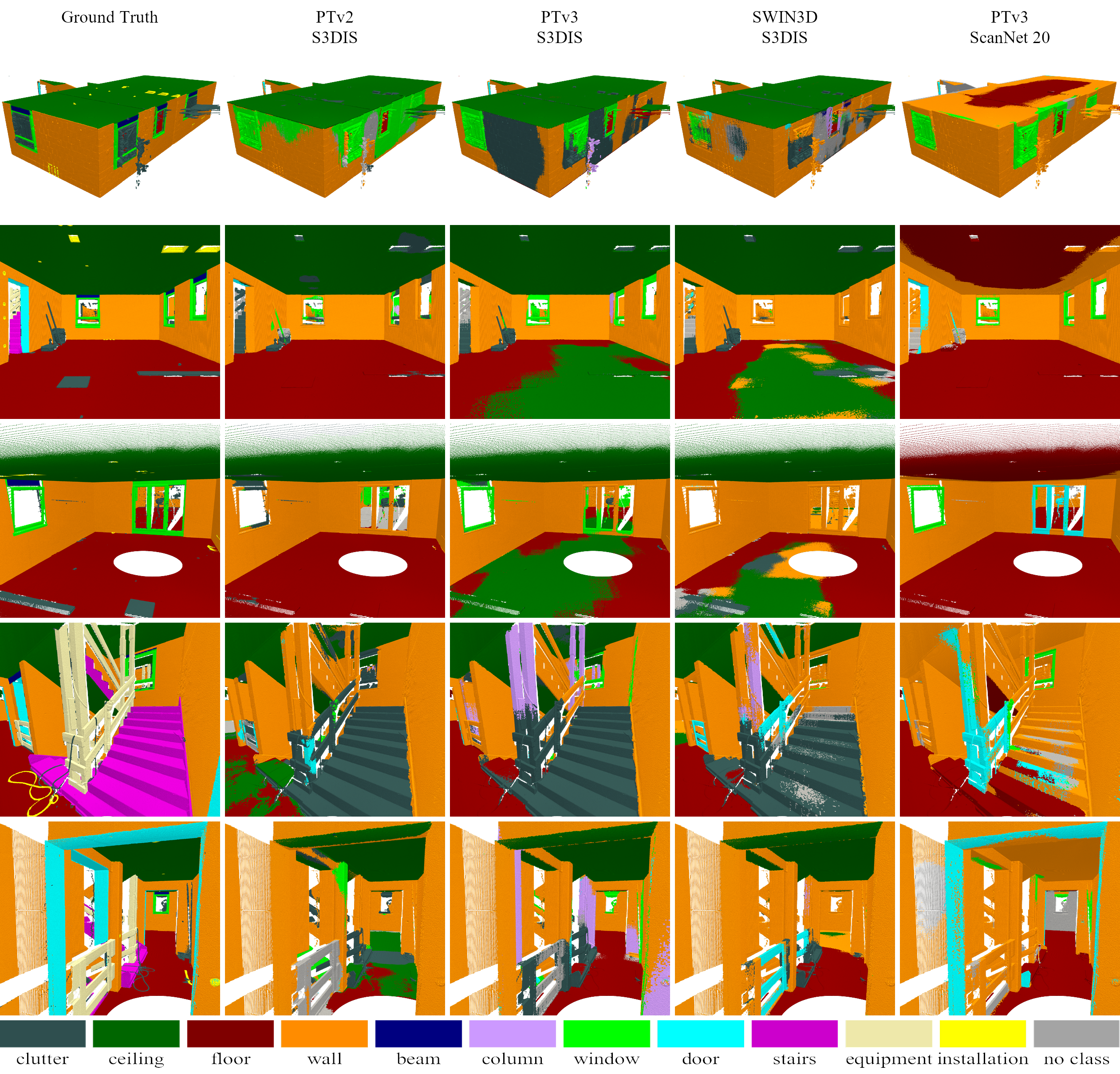}
  \caption{\label{fig:transfer_1} \textbf{Transfer Test Results for 3D Semantic Segmentation $I$.} 
  This figure presents the inference results from the transfer approach using three model architectures: Point Transformer V2, Point Transformer V3, and SWIN3D. Each model was trained on one of the two datasets, S3DIS or ScanNet 20, and tested on our custom validation dataset, which focuses on shell construction site scenes. The columns display five representative screenshots per model, with predicted class labels uniquely colored to illustrate the segmentation performance across different architectural components. Labels that cannot be translated to one of the 11 target classes, according to the encoding in Figure \ref{fig:class_encoding}, are assigned to an additional \textit{no class}.}
\end{figure*}

\begin{figure*}[]
  \centering
  \includegraphics[width=\textwidth]{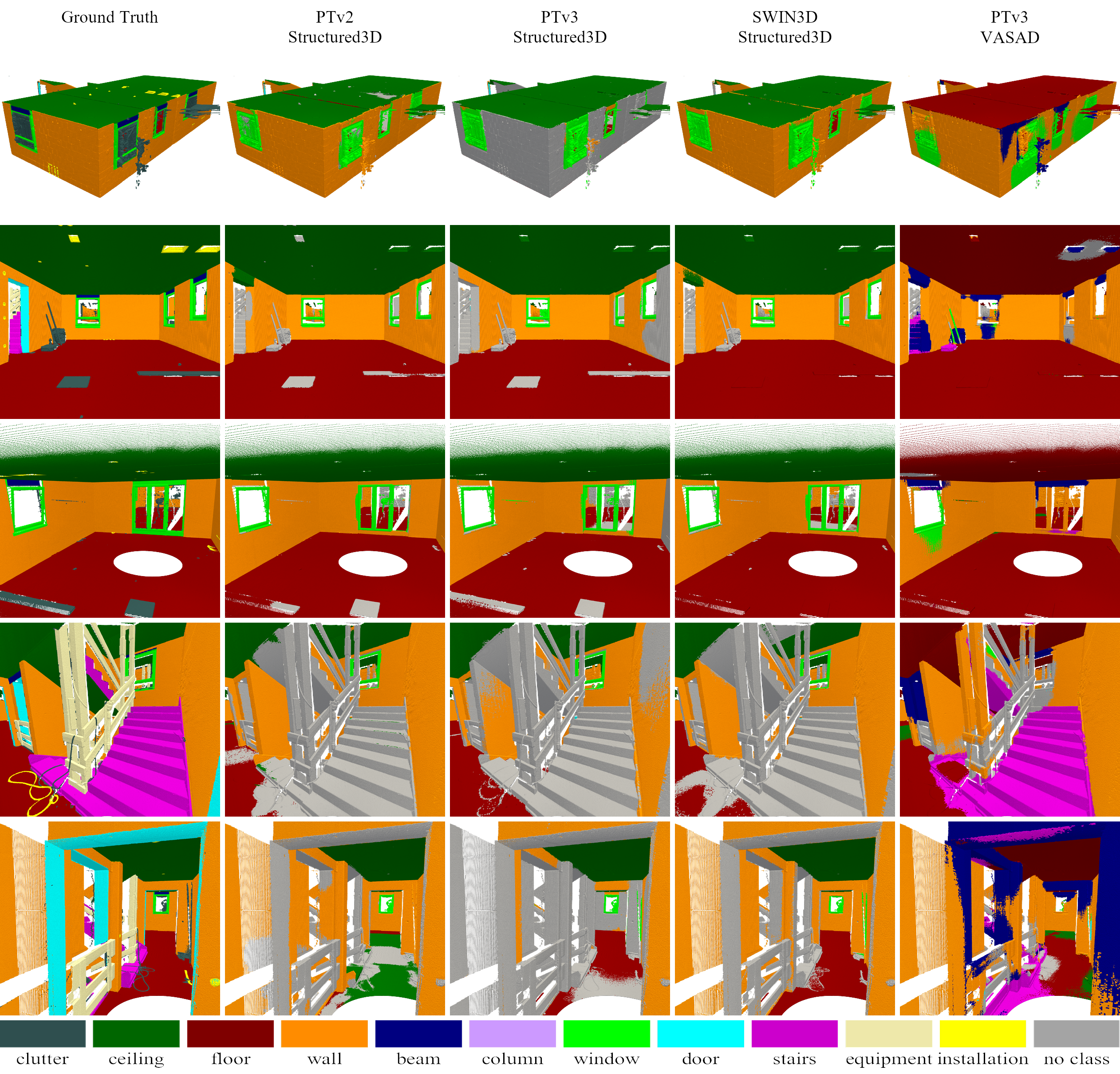}
  \caption{\label{fig:transfer_2} \textbf{Transfer Test Results for 3D Semantic Segmentation $II$.}
  This figure presents the inference results from the transfer approach using three model architectures: Point Transformer V2, Point Transformer V3, and SWIN3D. Each model was trained on one of the two datasets, Structured3D or VASAD, and tested on our custom validation dataset, which focuses on shell construction site scenes. The columns display five representative screenshots per model, with predicted class labels uniquely colored to illustrate the segmentation performance across different architectural components. Labels that cannot be translated to one of the 11 target classes, according to the encoding in Figure \ref{fig:class_encoding}, are assigned to an additional \textit{no class}.}
\end{figure*}

\def\tabcolwidth{.5cm}
\begin{table}[]
\begin{adjustwidth}{-1em}{-1em}
\small
\centering
    \caption{\textbf{Transfer Test Results for 3D Semantic Segmentation} 
    for three different model architectures — Point Transformer V2, Point Transformer V3, and SWIN3D — pre-trained on four open-source datasets for the task of 3D semantic segmentation. The performance metrics include class-wise Intersection over Union (IoU) and Accuracy (Acc), which were reported for validation during training. The test results marked with an asterisk (*) indicate the use of Test Time Augmentation (TTA) in the final testing to enhance prediction stability.}
    \label{tab:transfer}
    \centering
    \begin{tblr}{Q[l,f,1.3cm] 
      Q[c,f,\tabcolwidth] Q[c,f,\tabcolwidth] 
      Q[c,f,\tabcolwidth] Q[c,f,\tabcolwidth] 
      Q[c,f,\tabcolwidth] Q[c,f,\tabcolwidth]
      Q[c,f,\tabcolwidth] Q[c,f,\tabcolwidth]
      Q[c,f,\tabcolwidth] Q[c,f,\tabcolwidth]
      Q[c,f,\tabcolwidth] Q[c,f,\tabcolwidth]
      Q[c,f,\tabcolwidth] Q[c,f,\tabcolwidth]
      Q[c,f,\tabcolwidth] Q[c,f,\tabcolwidth]
      } 
         \SetCell[r=2]{f,l} {\textbf{target class}}
       & \SetCell[c=2]{f,c} {S3DIS\\(PTv2)} & 
       & \SetCell[c=2]{f,c} {S3DIS\\(PTv3)} & 
       & \SetCell[c=2]{f,c} {S3DIS\\(SWIN3D)} & 
       & \SetCell[c=2]{f,c} {ScanNet\\(PTv3)} & 
       & \SetCell[c=2]{f,c} {Structured3D\\(PTv2)} & 
       & \SetCell[c=2]{f,c} {Structured3D\\(PTv3)} & 
       & \SetCell[c=2]{f,c} {Structured3D\\(SWIN3D)} & 
       & \SetCell[c=2]{f,c} {VASAD\\(PTv3)} & 
       
       \\
      
      & \textbf{IoU*} & \textbf{Acc*} 
      & \textbf{IoU*} & \textbf{Acc*}
      & \textbf{IoU*} & \textbf{Acc*} 
      & \textbf{IoU*} & \textbf{Acc*}
      & \textbf{IoU*} & \textbf{Acc*} 
      & \textbf{IoU*} & \textbf{Acc*}
      & \textbf{IoU*} & \textbf{Acc*}
      & \textbf{IoU*} & \textbf{Acc*} \\
      
    \hline
    
    \_none       & .22 & .32 & .22 & .31 & .20 & .37 &										\\
    ceiling      & .93 & .99 & .76 & .83 & .96 & .99 &      &      & .95 & 1.0 & .99 & .99 & .93 & .93 & 0    & 0 \\
    floor        & .82 & .84 & .58 & .82 & .84 & .86 & .34 & 1.0 & .85 & .86 & .97 & .98 & .94 & .96 & .30 & .98 \\
    wall         & .86 & .92 & .86 & .93 & .69 & .81 & .60 & .82 & .94 & .99 & .44 & .45 & .89 & .92 & .65 & .69 \\
    beam         & 0    & 0    & 0    & 0    & 0    & 0    &      &      &      &      &      &      &      &      & 0	  & .45 \\
    column       & 0    & 0    & 0    & 0    & 0    & 0    &      &      &      &      &      &      &      &      &      &      \\
    window       & .13	& .33 & .20 & .29 & .03	& .03 & .28 & .61 & .28 & .43 & .29 & .48 & .29 & .50 & .14 & .42 \\
    door         & .06	& .08 & .15 & .22 & 0	& 0    & .08 & .68 & 0    & 0    & 0    & 0    & 0    & 0    & 0    & 0    \\
	stair        &      &      &      &      &      &      &      &      &      &      &      &      &      &      & .25 & .67 \\
    equipment    & \\
    installation & \\											
    
    \end{tblr}
\end{adjustwidth}
\end{table}

\subsection{Cross-Domain Results}

In the cross-domain experiment, the two baseline datasets, S3DIS and VASAD, had the widest overlap with our target classes. However, the results presented in Table \ref{tab:transfer} showed that this overlap was less relevant for the overall outcome than initially anticipated and went against the hypothesis. 
In this second experiment, the IoU metrics of the individual classes were compared with each other and not the averaged performance metrics to account for the fact that no baseline dataset contained all target classes. We also focused exclusively on classes that appeared in both the training and validation datasets. 
The IoU results for all classes across all models in Table \ref{tab:transfer} fell slightly to significantly below 30\%, making them unsuitable for pre-labeling applications, except for the three planar object classes: \textit{floor, ceiling}, and \textit{wall}, which were included in all datasets except for one case in ScanNet 20. The PTv3 model trained on the ScanNet 20 dataset showed promising segmentation results for \textit{floors}, but the dataset lacks a \textit{ceiling} class. Consequently, the model struggled with our real-world laser scans, frequently misinterpreting true ceilings in the validation data as \textit{walls} or \textit{floors}, as shown in Figure \ref{fig:transfer_1}. The only field where the model performed comparably to other models was in the segmentation of \textit{windows}. Therefore, we excluded the PTv3 model trained on ScanNet from our subsequent fine-tuning experiments.
The PTv3 model trained on the VASAD dataset could also be excluded from our subsequent experiments, based on the results in Table \ref{tab:transfer} and Figure \ref{fig:transfer_2}. The primary issue was the confusion between the three classes of \textit{wall, ceiling}, and \textit{floor} in the target dataset, which corresponded to the four classes of \textit{partition wall, bearing wall, slab}, and \textit{ceiling} in the VASAD dataset. This confusion caused the model to fail to recognize ceilings and to oscillate uncertainly between \textit{partition wall} and \textit{bearing wall} for \textit{walls}. 
The visual evaluation also showed strong smearing of the classes \textit{beam} and \textit{window} into the \textit{wall} segments. 
VASAD was the only one of the four baseline datasets that included \textit{stairs} as a distinct class and could roughly detect them within the point cloud. Still, the segmentation was highly inaccurate and insufficient for practical application. By contrast, models trained with datasets that did not contain a \textit{stair} class couldn't assign a dedicated label to the \textit{stair} segments but did more consistently isolate the stairs with clean edges from the environment. We suspect the poor results are directly related to the domain gap between the characteristics of imperfect real-world scan data and synthetic training data. The same domain gap existed with the synthetic Structured3D dataset, but Structured3D is significantly larger and more diverse, which helped to mitigate the synthetic data discrepancy.

The evaluation of the two Point Transformer models, PTv2 and PTv3, along with the SWIN3D model trained on the S3DIS dataset, showed ambivalent results when attempting to transfer what has been learned to our validation dataset. Ceilings were uniformly well-segmented, though the PTv3 model fell quantitatively behind the performance of the two competing models in Table \ref{tab:transfer}. Floors and walls posed significant challenges for all three models, as strikingly illustrated in Figure \ref{fig:transfer_1}.
The most promising result in the second experimental series came from the SWIN3D model trained on the synthetic Structured3D dataset. Although the model had never seen data from our validation dataset during training, its inference on real laser scanner data yielded an IoU of over 89\% for \textit{ceilings, walls}, and \textit{floors}. Despite the Structured3D dataset having the smallest overlap with our target classes, the consistent assignment of unrecognized objects to an additional \textit{no\_class} group could facilitate post-processing when creating annotations for a new dataset. The PTv2 model delivered comparable results but was slightly less accurate in the qualitative segmentation of floor surfaces. The PTv3 model, however, struggled significantly with the correct assignment of \textit{wall} elements, as shown in Figure \ref{fig:transfer_2}.

The results of the cross-domain evaluation experiments demonstrated that, despite the apparent domain gap between the training and testing data, most models managed to perceive the scene content in principle. The predictions still appeared as though they were from a model trained only halfway or with insufficient data. Still, an essential synergy was recognizable between the residential indoor datasets and our target dataset, collected at an early-stage shell construction site and not in a finished and furnished building. This synergy could be further leveraged in a transfer learning approach for object segmentation across fundamentally different domains. 

\def\tabcolwidth{.7cm}
\begin{table*}[]
    \caption{\textbf{Transfer-learning Validation and Test Results for 3D Semantic Segmentation}
     for two model architectures — Point Transformer V3 and SWIN3D — pre-trained with two open-source datasets and fine-tuned on our validation dataset for the task of 3D semantic segmentation. The performance metrics include mean Intersection over Union (mIoU), mean Accuracy (mAcc), and overall Accuracy (allAcc), which are reported for both validation during training and final testing. The test results marked with an asterisk (*) indicate the use of Test Time Augmentation (TTA) in the final testing to enhance prediction stability.}
    \label{tab:finetune}
    \centering
    \begin{tblr}{
     Q[l,f,1.7cm] Q[l,f,1.2cm] Q[l,f,1cm] Q[c,f,1.2cm]
     Q[c,f,\tabcolwidth] Q[c,f,\tabcolwidth] Q[c,f,\tabcolwidth] 
     Q[c,f,\tabcolwidth] Q[c,f,\tabcolwidth] Q[c,f,\tabcolwidth] 
     } 

    \SetCell[c=2]{f,c} {Dataset} \\
    
    \textbf{Pretrain} & \textbf{Finetune} & \textbf{Archit.} & \textbf{num classes} & \textbf{mIoU} & \textbf{mAcc} & \textbf{allAcc} & 
    \textbf{mIoU*} & \textbf{mAcc*} & \textbf{allAcc*} \\
    \hline

    S3DIS        & Ours & PTv3   & 11 & .59 & .66 & .95 & .73 & .81 & .99 \\
    S3DIS        & Ours & SWIN3D & 11 & .64 & .71 & .95 & .73 & .81 & .99 \\
    Structured3D & Ours & PTv3   & 11 & .68	& .76 & .96 & .75 & .83 & .99 \\
    Structured3D & Ours & SWIN3D & 11 & .65 & .73 & .96 & .77 & .83 & .99 \\

    \end{tblr}
\end{table*}

\begin{figure*}[]
  \centering
  \includegraphics[width=\textwidth]{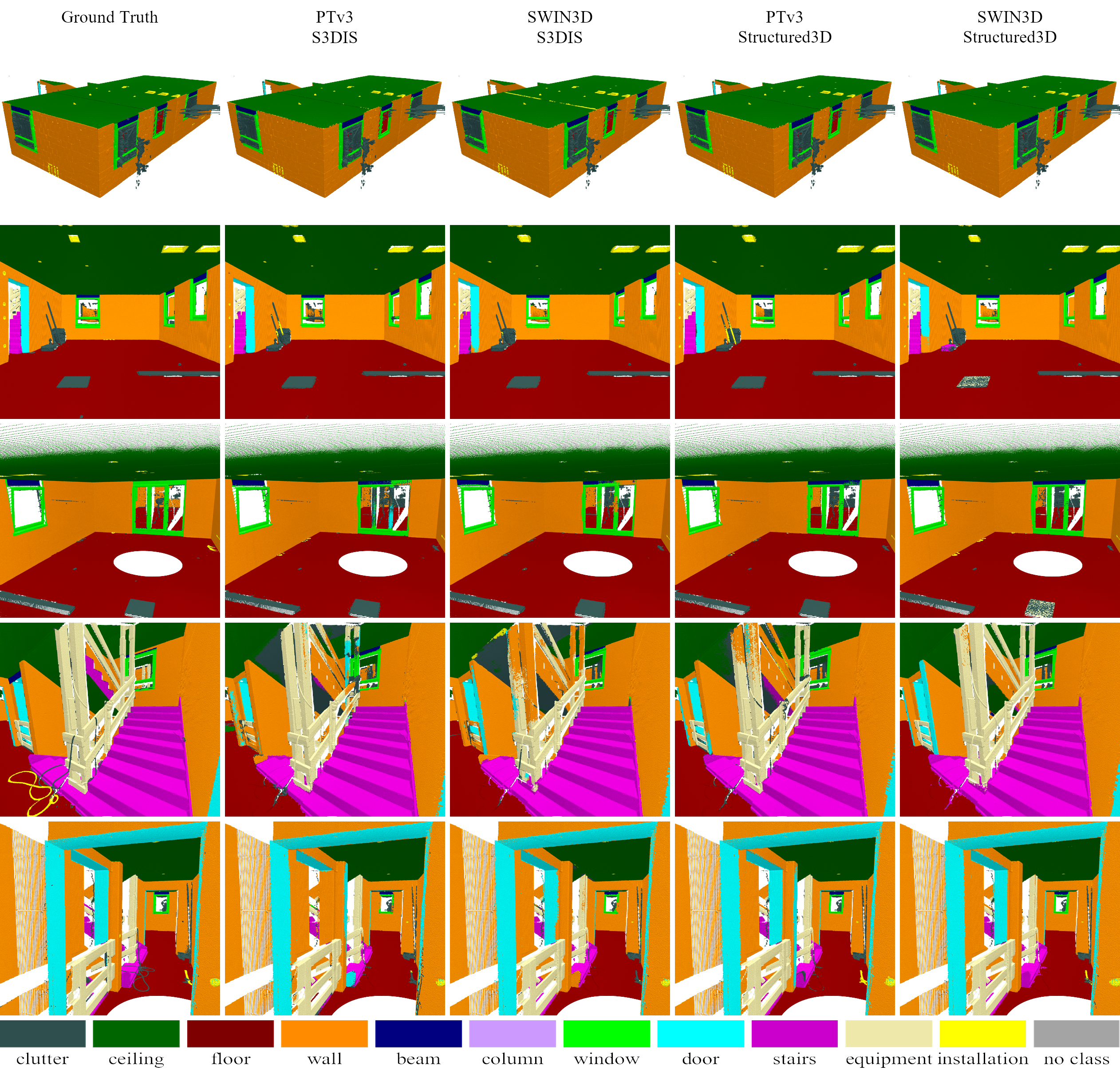}
  \caption{\label{fig:finetune} \textbf{Transfer-Learning Test Results for 3D Semantic Segmentation.}
  This figure presents the inference results from the transfer-learning experiment using two model architectures: Point Transformer V3, and SWIN3D. Each model was pre-trained on one of the two datasets, S3DIS or Structured3D, then further fine-tuned and tested on our custom validation dataset, which focuses on shell construction site scenes. The columns display five representative screenshots per model, with predicted class labels uniquely colored to illustrate the segmentation performance across different architectural components.}
\end{figure*}

\subsection{Transfer Learning Results}

In the final section, we evaluated the results of the transfer learning approach. The two model architectures, PTv3 and SWIN3D, were pre-trained on the two most promising datasets, S3DIS and Structured3D, and then fine-tuned for the target classes of our validation dataset. The results are presented in Table \ref{tab:finetune} and Figure \ref{fig:finetune}.
As expected, the precise test results were about 10\% higher than the validation results during fine-tuning, which could be attributed to Test-Time Augmentation \cite{kimura.Understanding.2021}. Notably, the results of the four combinations in Table \ref{tab:finetune} were all very close. On average, all models achieved a segmentation mIoU of approximately 75\% ($\pm$2\%) after fine-tuning. These results were further confirmed by the qualitative evaluation in Figure \ref{fig:finetune}. The comparison showed that the predictions of all four combinations were very similar to the ground truth segments, differing only in fine details. All models consistently segmented \textit{ceilings, walls}, and \textit{floors} with minor oscillations at the edges between adjacent planes. The predictions were generated on a voxelized subsample of the point cloud, though the oscillations most likely arose from voxel-based upscaling rather than mistaken model predictions. \textit{Beam} elements, such as those found over window lintels, were roughly recognized, but the models often struggled to distinguish them from surrounding components clearly. Columns do not appear in our test dataset, as they are uncommon in apartment buildings and thus couldn't be further evaluated here.
In segmenting \textit{window} objects, the two SWIN3D models showed a relative advantage over the two PTv3 models. The same held true for \textit{door} frames. Generally, the models pre-trained on the Structured3D dataset were more accurate in absolute terms. In segmenting \textit{stairs}, the PTv3 models had an advantage in our experiments. We also found significantly better results in segmenting small electrical \textit{installations} when using pre-training with the synthetic but significantly larger Structured3D dataset. Overall, the best results in our three experiments were achieved through the transfer learning approach with the Structured3D dataset, which we fine-tuned with a small amount of labeled training data from the target domain.

\section{Limitations}

While this study provides valuable insights into the application of existing 3D semantic segmentation models within the context of shell construction sites, and the outcomes met the expectations of this test series, a few limitations should be acknowledged.

The results of this paper are proof of concept and do not replace a full-scale study, which would require a diverse validation dataset with a wide variety of representative construction sites but also edge cases that rarely occur. The authors recognize that our custom dataset is small compared to standard deep learning training sets. All laser scans were recorded in a single building, leading to a lack of general diversity. We avoid making assumptions about how well the fine-tuned models would generalize to other building classes and construction types. Consequently, our results can only be applied in scenarios where the model is deployed on point clouds from similar buildings. 


We acknowledge that utilizing normal vectors as additional point features should further improve segmentation results, and we will continue our future research in this area. However, to allow for a fair comparison across different datasets in this test series, the input feature set for all models had to be reduced to the lowest common denominator. Normal vectors can be natively extracted from the CAD model for synthetic training data. For real-world laser scans, the normal vector estimation depends on the choice of appropriate algorithms, which can introduce significant errors as they all have their flaws. The same applies to the scalar intensity of surface light reflection from LiDAR scanners, which might provide helpful information for object segmentation. This feature is almost impossible to simulate; thus, none of the baseline datasets include reflection intensity. Since RGB color was the point feature most datasets included, we decided to limit the experimental design to color features only, even though our results fell short of the best benchmark results that utilized normals.

\section{Conclusion}

This study explored the potential of adapting existing datasets and model architectures for 3D point cloud semantic segmentation in the industry-specific context of shell construction sites. Our work was motivated by the need to repurpose the capabilities of existing segmentation approaches, predominantly developed for interior perception in indoor spaces, to more technical and structurally complex environments within the Architecture, Engineering, and Construction (AEC) domains.

We established a baseline by training and evaluating three state-of-the-art model architectures — Point Transformer V2, Point Transformer V3, and SWIN3D — on our custom validation dataset. The results demonstrated that these models could achieve promising segmentation results even with limited domain-specific data, although challenges remain in accurately segmenting more complex structural components.

Building on these findings, we applied a straightforward transfer inference approach to evaluate the generalization of models pre-trained on large-scale public datasets — S3DIS, ScanNet 20, Structured3D, and VASAD — when applied to our shell construction dataset. Despite the inherent domain shift, the models demonstrated reasonable performance, underscoring the potential of transfer learning to reduce the need for extensive labeled data in new and specialized domains. However, certain classes, particularly those unique to construction sites, proved more challenging to segment accurately, emphasizing the critical need for high-quality domain-specific data.

Ensemble learning strategies were not investigated in this study. Stacking multiple base models to obtain final predictions with better performance is a popular method to boost underrepresented classes, and we identified potential for ensemble learning in our experiments where certain models performed best on specific classes. However, this was not the primary focus of this work and should be extensively investigated in detail in further studies. 

Finally, transfer learning showed that with a small amount of labeled data, the pre-trained baseline models could further improve their segmentation performance, particularly the capability to distinguish between similar architectural components. This supports our thesis that established datasets for point cloud semantic segmentation can facilitate data annotation by automating the pre-labeling process, even if the base data does not perfectly match the target data.

Overall, our experiments confirm that combining pre-trained transformer architectures with minimal fine-tuning on domain-specific data can be an effective strategy for semantic segmentation in the AEC industry. However, progress in this area will require high-quality datasets with domain-specific expertise. Even if approaches like transfer learning, unsupervised learning, or 3D Segment Anything \cite{yang.SAM3D.2023} minimizes the need for extensive training data; appropriate benchmark datasets are still necessary for validation. 

Future work should focus on expanding the availability of data resources, but universities and independent research institutions alone cannot achieve this. The stakeholders who will benefit most from the development are the large construction groups, construction project developers, and public authorities. They are, therefore, responsible. Datasets need to be open-sourced to enable collaboration. Many industries have already internalized this, resulting in a clear lead over the construction industry. Our following contributions will emphasize creating a larger AEC-specific point cloud perception dataset based on the results of this work and our experience building the validation data set used here. While state-of-the-art semantic segmentation of point clouds has made significant progress, future milestones should include instance segmentation and material-based classification.


\printbibliography

@article{armeni.3D.2016,
  title = {{{3D Semantic Parsing}} of {{Large-Scale Indoor Spaces}}},
  author = {Armeni, Iro and Sener, Ozan and Zamir, Amir Roshan and Jiang, Helen and Brilakis, Ioannis and Fischer, Martin and Savarese, Silvio},
  year = {2016},
  month = jun,
  pages = {1534--1543},
  doi = {10.1109/cvpr.2016.170}
}

@article{guo.Deep.2021,
  title = {Deep {{Learning}} for {{3D Point Clouds}}: {{A Survey}}},
  shorttitle = {Deep {{Learning}} for {{3D Point Clouds}}},
  author = {Guo, Yulan and Wang, Hanyun and Hu, Qingyong and Liu, Hao and Liu, Li and Bennamoun, Mohammed},
  year = {2021},
  month = dec,
  journal = {IEEE Transactions on Pattern Analysis and Machine Intelligence},
  volume = {43},
  number = {12},
  pages = {4338--4364},
  issn = {0162-8828, 2160-9292, 1939-3539},
  doi = {10.1109/TPAMI.2020.3005434}
}

@incollection{zheng.Structured3D.2020,
  title = {{{Structured3D}}: {{A Large Photo-Realistic Dataset}} for {{Structured 3D Modeling}}},
  shorttitle = {{{Structured3D}}},
  booktitle = {Computer {{Vision}} -- {{ECCV}} 2020},
  author = {Zheng, Jia and Zhang, Junfei and Li, Jing and Tang, Rui and Gao, Shenghua and Zhou, Zihan},
  editor = {Vedaldi, Andrea and Bischof, Horst and Brox, Thomas and Frahm, Jan-Michael},
  year = {2020},
  volume = {12354},
  pages = {519--535},
  publisher = {Springer International Publishing},
  address = {Cham},
  doi = {10.1007/978-3-030-58545-7\_30},
  urldate = {2024-03-18},
  isbn = {978-3-030-58544-0 978-3-030-58545-7    },
  langid = {english}
}

@article{noichl.Enhancing.2024,
  title = {Enhancing Point Cloud Semantic Segmentation in the Data-scarce Domain of Industrial Plants through Synthetic Data},
  author = {Noichl, Florian and Collins, Fiona C. and Braun, Alexander and Borrmann, Andr{\'e}},
  year = {2024},
  month = jan,
  journal = {Computer-Aided Civil and Infrastructure Engineering},
  pages = {mice.13153},
  issn = {1093-9687, 1467-8667},
  doi = {10.1111/mice.13153},
  langid = {english}
}

@article{yang.SAM3D.2023,
  title = {{{SAM3D}}: {{Segment Anything}} in {{3D Scenes}}},
  shorttitle = {{{SAM3D}}},
  author = {Yang, Yunhan and Wu, Xiaoyang and He, Tong and Zhao, Hengshuang and Liu, Xihui},
  year = {2023},
  publisher = {[object Object]},
  doi = {10.48550/ARXIV.2306.03908},
  urldate = {2024-03-18},
  copyright = {arXiv.org perpetual, non-exclusive license}
}

@article{wu.Largescale.2023,
  title = {Towards {{Large-scale 3D Representation Learning}} with {{Multi-dataset Point Prompt Training}}},
  author = {Wu, Xiaoyang and Tian, Zhuotao and Wen, Xin and Peng, Bohao and Liu, Xihui and Yu, Kaicheng and Zhao, Hengshuang},
  year = {2023},
  publisher = {[object Object]},
  doi = {10.48550/ARXIV.2308.09718},
  copyright = {arXiv.org perpetual, non-exclusive license}
}

@article{kim.Learning.2022,
  title = {Learning {{Semantic Segmentation}} from {{Multiple Datasets}} with {{Label Shifts}}},
  author = {Kim, Dongwan and Tsai, Yi-Hsuan and Suh, Yumin and Faraki, Masoud and Garg, Sparsh and Chandraker, Manmohan and Han, Bohyung},
  year = {2022},
  publisher = {[object Object]},
  doi = {10.48550/ARXIV.2202.14030},
  copyright = {arXiv.org perpetual, non-exclusive license}
}

@article{wang.CrossDataset.2021,
  title = {Cross-{{Dataset Collaborative Learning}} for {{Semantic Segmentation}} in {{Autonomous Driving}}},
  author = {Wang, Li and Li, Dong and Liu, Han and Peng, Jinzhang and Tian, Lu and Shan, Yi},
  year = {2021},
  publisher = {[object Object]},
  doi = {10.48550/ARXIV.2103.11351},
  copyright = {arXiv.org perpetual, non-exclusive license}
}

@misc{pointceptcontributors.Pointcept.2023,
  title = {Pointcept: {{A Codebase}} for {{Point Cloud Perception Research}} {, https://github.com/Pointcept/Pointcept}},
  author = {Pointcept Contributors},
  year = {2023},
  urldate = {2024-03-16}
}

@inproceedings{dai.ScanNet.2017a,
  title = {{{ScanNet}}: {{Richly-Annotated 3D Reconstructions}} of {{Indoor Scenes}}},
  shorttitle = {{{ScanNet}}},
  booktitle = {2017 {{IEEE Conference}} on {{Computer Vision}} and {{Pattern Recognition}} ({{CVPR}})},
  author = {Dai, Angela and Chang, Angel X. and Savva, Manolis and Halber, Maciej and Funkhouser, Thomas and Niessner, Matthias},
  year = {2017},
  month = jul,
  pages = {2432--2443},
  publisher = {IEEE},
  address = {Honolulu, HI},
  doi = {10.1109/CVPR.2017.261},
  urldate = {2024-03-18},
  isbn = {978-1-5386-0457-1}
}

@inproceedings{langlois.VASAD.2022,
  title = {{{VASAD}}: A {{Volume}} and {{Semantic}} Dataset for {{Building Reconstruction}} from {{Point Clouds}}},
  shorttitle = {{{VASAD}}},
  booktitle = {2022 26th {{International Conference}} on {{Pattern Recognition}} ({{ICPR}})},
  author = {Langlois, Pierre-Alain and Xiao, Yang and Boulch, Alexandre and Marlet, Renaud},
  year = {2022},
  month = aug,
  pages = {4008--4015},
  publisher = {IEEE},
  address = {Montreal, QC, Canada},
  doi = {10.1109/ICPR56361.2022.9956356},
  isbn = {978-1-66549-062-7}
}

@inproceedings{zhao.Point.2021,
  title = {Point {{Transformer}}},
  booktitle = {{{IEEE International Conference}} on {{Computer Vision}} ({{ICCV}}) 2021},
  author = {Zhao, Hengshuang and Jiang, Li and Jia, Jiaya and Torr, Philip and Koltun, Vladlen},
  year = {2021},
  month = sep,
  address = {Virtual},
  doi = {10.48550/arXiv.2012.09164},
  urldate = {2024-03-18}
}

@inproceedings{wu.Point.2022,
  title = {Point {{Transformer V2}}: {{Grouped Vector Attention}} and {{Partition-based Pooling}}},
  shorttitle = {Point {{Transformer V2}}},
  booktitle = {Conference on {{Neural Information Processing Systems}} ({{NeurIPS}}) 2022},
  author = {Wu, Xiaoyang and Lao, Yixing and Jiang, Li and Liu, Xihui and Zhao, Hengshuang},
  year = {2022},
  month = oct,
  eprint = {2210.05666},
  primaryclass = {cs},
  doi = {10.48550/arXiv.2210.05666},
  archiveprefix = {arXiv}
}

@inproceedings{wu.Point.2023,
  title = {Point {{Transformer V3}}: {{Simpler}}, {{Faster}}, {{Stronger}}},
  shorttitle = {Point {{Transformer V3}}},
  booktitle = {{{IEEE Conference}} on {{Computer Vision}} and {{Pattern Recognition}} ({{CVPR}}) 2024},
  author = {Wu, Xiaoyang and Jiang, Li and Wang, Peng-Shuai and Liu, Zhijian and Liu, Xihui and Qiao, Yu and Ouyang, Wanli and He, Tong and Zhao, Hengshuang},
  year = {2023},
  month = dec,
  eprint = {2312.10035},
  primaryclass = {cs},
  publisher = {arXiv},
  doi = {10.48550/arXiv.2312.10035},
  urldate = {2024-03-18},
  archiveprefix = {arXiv}
}

@misc{yang.Swin3D.2023,
  title = {{{Swin3D}}: {{A Pretrained Transformer Backbone}} for {{3D Indoor Scene Understanding}}},
  shorttitle = {{{Swin3D}}},
  author = {Yang, Yu-Qi and Guo, Yu-Xiao and Xiong, Jian-Yu and Liu, Yang and Pan, Hao and Wang, Peng-Shuai and Tong, Xin and Guo, Baining},
  year = {2023},
  month = aug,
  number = {arXiv:2304.06906}   ,
  eprint = {2304.06906},
  primaryclass = {cs},
  publisher = {arXiv},
  doi = {10.48550/arXiv.2304.06906},
  archiveprefix = {arXiv}
}

@misc{paperswithcode.3D.2024,
  title = {{{3D Semantic Segmentation Benchmarks} {https://paperswithcode.com/task/3d-semantic-segmentation/codeless}}},
  author = {{PapersWithCode.com}},
  year = {2024},
  month = mar,
  urldate = {2024-03-16}
}

@article{rauch.Semantic.2023,
  title = {Semantic {{Point Cloud Segmentation}} with {{Deep-Learning-Based Approaches}} for the {{Construction Industry}}: {{A Survey}}},
  shorttitle = {Semantic {{Point Cloud Segmentation}} with {{Deep-Learning-Based Approaches}} for the {{Construction Industry}}},
  author = {Rauch, Lukas and Braml, Thomas},
  year = {2023},
  month = aug,
  journal = {Applied Sciences},
  volume = {13},
  number = {16},
  pages = {9146},
  issn = {2076-3417},
  doi = {10.3390/app13169146},
  langid = {english}
}

@article{dimitrov.Segmentation.2015,
  title = {Segmentation of Building Point Cloud Models Including Detailed Architectural/Structural Features and {{MEP}} Systems},
  author = {Dimitrov, Andrey and {Golparvar-Fard}, Mani},
  year = {2015},
  month = mar,
  journal = {Automation in Construction},
  volume = {51},
  pages = {32--45},
  issn = {09265805},
  doi = {10.1016/j.autcon.2014.12.015},
  langid = {english}
}

@article{yin.Automated.2021a,
  title = {Automated Semantic Segmentation of Industrial Point Clouds Using {{ResPointNet}}++},
  author = {Yin, Chao and Wang, Boyu and Gan, Vincent J.L. and Wang, Mingzhu and Cheng, Jack C.P.},
  year = {2021},
  month = oct,
  journal = {Automation in Construction},
  volume = {130},
  pages = {103874},
  issn = {09265805},
  doi = {10.1016/j.autcon.2021.103874},
  urldate = {2024-03-18},
  langid = {english}
}

@article{perez-perez.Segmentation.2021a,
  title = {Segmentation of Point Clouds via Joint Semantic and Geometric Features for {{3D}} Modeling of the Built Environment},
  author = {{Perez-Perez}, Yeritza and {Golparvar-Fard}, Mani and {El-Rayes}, Khaled},
  year = {2021},
  month = may,
  journal = {Automation in Construction},
  volume = {125},
  pages = {103584},
  issn = {09265805},
  doi = {10.1016/j.autcon.2021.103584},
  urldate = {2024-03-18},
  langid = {english}
}

@article{ma.Semantic.2020a,
  title = {Semantic Segmentation of Point Clouds of Building Interiors with Deep Learning: {{Augmenting}} Training Datasets with Synthetic {{BIM-based}} Point Clouds},
  shorttitle = {Semantic Segmentation of Point Clouds of Building Interiors with Deep Learning},
  author = {Ma, Jong Won and Czerniawski, Thomas and Leite, Fernanda},
  year = {2020},
  month = may,
  journal = {Automation in Construction},
  volume = {113},
  pages = {103144},
  issn = {09265805},
  doi = {10.1016/j.autcon.2020.103144   },
  urldate = {2024-03-18},
  langid = {english}
}

@article{guo.PCT.2021,
  title = {{{PCT}}: {{Point}} Cloud Transformer},
  shorttitle = {{{PCT}}},
  author = {Guo, Meng-Hao and Cai, Jun-Xiong and Liu, Zheng-Ning and Mu, Tai-Jiang and Martin, Ralph R. and Hu, Shi-Min},
  year = {2021},
  month = jun,
  journal = {Computational Visual Media},
  volume = {7},
  number = {2},
  pages = {187--199},
  issn = {2096-0433, 2096-0662},
  doi = {10.1007/s41095-021-0229-5},
  urldate = {2024-05-14},
  langid = {english}
}

@article{liu.Swin.2021,
  title = {Swin {{Transformer}}: {{Hierarchical Vision Transformer}} Using {{Shifted Windows}}},
  shorttitle = {Swin {{Transformer}}},
  author = {Liu, Ze and Lin, Yutong and Cao, Yue and Hu, Han and Wei, Yixuan and Zhang, Zheng and Lin, Stephen and Guo, Baining},
  year = {2021},
  publisher = {[object Object]},
  doi = {10.48550/ARXIV.2103.14030},
  urldate = {2024-05-14},
  copyright = {Creative Commons Attribution 4.0 International}
}

@article{lai.Stratified.2022,
  title = {Stratified {{Transformer}} for {{3D Point Cloud Segmentation}}},
  author = {Lai, Xin and Liu, Jianhui and Jiang, Li and Wang, Liwei and Zhao, Hengshuang and Liu, Shu and Qi, Xiaojuan and Jia, Jiaya},
  year = {2022},
  publisher = {[object Object]},
  doi = {10.48550/ARXIV.2203.14508},
  urldate = {2024-05-14},
  copyright = {Creative Commons Attribution Non Commercial No Derivatives 4.0 International}
}

@article{bao.BEiT.2021,
  title = {{{BEiT}}: {{BERT Pre-Training}} of {{Image Transformers}}},
  shorttitle = {{{BEiT}}},
  author = {Bao, Hangbo and Dong, Li and Piao, Songhao and Wei, Furu},
  year = {2021},
  publisher = {[object Object]},
  doi = {10.48550/ARXIV.2106.08254},
  urldate = {2024-05-14},
  copyright = {arXiv.org perpetual, non-exclusive license}
}

@article{devlin.BERT.2018,
  title = {{{BERT}}: {{Pre-training}} of {{Deep Bidirectional Transformers}} for {{Language Understanding}}},
  shorttitle = {{{BERT}}},
  author = {Devlin, Jacob and Chang, Ming-Wei and Lee, Kenton and Toutanova, Kristina},
  year = {2018},
  publisher = {[object Object]},
  doi = {10.48550/ARXIV.1810.04805},
  urldate = {2024-05-14},
  copyright = {arXiv.org perpetual, non-exclusive license}
}

@misc{berman.LovaszSoftmax.2018,
  title = {The {{Lovasz-Softmax}} Loss: {{A}} Tractable Surrogate for the Optimization of the Intersection-over-Union Measure in Neural Networks},
  shorttitle = {The {{Lov}}{\textbackslash}'asz-{{Softmax}} Loss},
  author = {Berman, Maxim and Triki, Amal Rannen and Blaschko, Matthew B.},
  year = {2018},
  month = apr,
  number = {arXiv:1705.08790},
  eprint = {1705.08790},
  primaryclass = {cs},
  publisher = {arXiv},
  urldate = {2024-07-15},
  archiveprefix = {arXiv}
}

@misc{qi.PointNet.2016a,
  title = {{{PointNet}}: {{Deep Learning}} on {{Point Sets}} for {{3D Classification}} and {{Segmentation}}},
  shorttitle = {{{PointNet}}},
  author = {Qi, Charles R. and Su, Hao and Mo, Kaichun and Guibas, Leonidas J.},
  year = {2016},
  publisher = {arXiv},
  doi = {10.48550/ARXIV.1612.00593},
  urldate = {2024-08-26},
  copyright = {arXiv.org perpetual, non-exclusive license}
}

@incollection{kimura.Understanding.2021,
  title = {Understanding {{Test-Time Augmentation}}},
  booktitle = {Neural {{Information Processing}}},
  author = {Kimura, Masanari},
  editor = {Mantoro, Teddy and Lee, Minho and Ayu, Media Anugerah and Wong, Kok Wai and Hidayanto, Achmad Nizar},
  year = {2021},
  volume = {13108},
  pages = {558--569},
  publisher = {Springer International Publishing},
  address = {Cham},
  doi = {10.1007/978-3-030-92185-9\_46},
  urldate = {2024-08-27},
  isbn = {978-3-030-92184-2 978-3-030-92185-9},
  langid = {english}
}






\end{document}